\documentclass{article}

\usepackage{graphicx}
\usepackage{afterpage}
\usepackage{arxiv}
\usepackage[title]{appendix}

\usepackage[utf8]{inputenc} % allow utf-8 input
\usepackage[T1]{fontenc}    % use 8-bit T1 fonts
\usepackage{hyperref}       % hyperlinks
\usepackage{url}            % simple URL typesetting
\usepackage{booktabs}       % professional-quality tables
\usepackage{amsfonts}       % blackboard math symbols
\usepackage{nicefrac}       % compact symbols for 1/2, etc.
\usepackage{microtype}      % microtypography
\usepackage{lipsum}		% Can be removed after putting your text content

\title{Training neural networks to encode symbols enables combinatorial generalization}

\date{} 					% Or removing it

\author{
  Ivan I. Vankov \\
  Department of Cognitive Science and Psychology\\
  New Bulgarian University\\  
  \texttt{i.i.vankov@cogs.nbu.bg} \\
   \And
 Jeffrey S. Bowers\\
  School of Psychological Science\\
  University of Bristol\\
  \texttt{j.bowers@bristol.ac.uk } \\
}

\begin{document}
Vankov, I., \& Bowers, J. (in press). Training neural networks to encode symbols enables combinatorial generalization. {\em Philosophical Transactions of the Royal Society B}. doi: 10.1098/rstb.2019.0309

\maketitle

\begin{abstract}
Combinatorial generalization - the ability to understand and produce novel combinations of already familiar elements - is considered to be a core capacity of the human mind and a major challenge to neural network models. A significant body of research suggests that conventional neural networks can’t solve this problem unless they are endowed with mechanisms specifically engineered for the purpose of representing symbols. In this paper we introduce a novel way of representing symbolic structures in connectionist terms - the vectors approach to representing symbols (VARS), which allows training standard neural architectures to encode symbolic knowledge explicitly at their output layers. In two simulations , we show that neural networks not only can learn to produce VARS representations, but in doing so they achieve combinatorial generalization in their symbolic and non-symbolic output. This adds to other recent work that has shown improved combinatorial generalization under specific training conditions, and raises the question of whether specific mechanisms or training routines are needed to support symbolic processing.
\end{abstract}

% keywords can be removed
\keywords{combinatorial generalization, symbolic processing, neural networks}

\section{Introduction} Recent advances in neural network modeling have led to impressive results in fields as diverse as object, face and scene recognition (LeCun, Bengio \& Hinton, 2015), reasoning (Silver et al, 2016), speech perception (Graves, Mohamed \& Hinton. 2013), machine translation (Wu et al), playing computer games (Antonoglou et al, 2015) and producing art (Gatys, Ecker \& Bethge. 2015). These successes have relied on a restricted set of tools (e.g., the back propagation learning algorithm or the convolutional network architecture), and principles (all learning and computations take place in links between units), and are consistent with the claim that all forms of cognition rely on a small set of general mechanisms in which minimal innate structure needs to be included.  We will call this class of connectionist models currently popular in computer science the “conventional connectionist framework”.
 \footnote{An earlier generation of models that adopted this approach has been characterized as “eliminative connectionism” (Pinker \& Prince, 1988) or “non-symbolic connectionism” (Holyoak \& Hummel, 2000).}. However there are still fundamental disagreements concerning the limitations of this framework, with some researchers claiming that it cannot account for a range of core cognitive capacities (e.g., Bowers, 2017; Marcus, 2018).

One of the main criticisms of the conventional connectionist approach is that it lacks the ability to represent symbolic structures and is thus unsuitable for modeling tasks requiring symbolic operations (e.g., Fodor \& Pylyshyn, 1988). It is important to emphasize that the critics of this approach are not claiming that connectionist systems are in principle unable to account for symbolic processing, but rather, that models need to be augmented in order to explicitly implement symbolic computation (Holyoak \& Hummel, 2000). This can involve introducing new computational mechanisms above and beyond the modification of weights between units, such the synchrony of units firing (Hummel \& Biederman, 1992; Hummel, \& Holyoak, 1997; Holyoak \& Hummel, 2003; Doumas, Hummel, \& Sandhofer, 2008) or the reliance of delay lines (Davis, 2010), introducing new inductive biases specifically designed to support symbolic computation, such as graph based structures (Battaglia et al., 2018), hybrid symbolic-connectionist units (Petrov \& Kokinov, 2001) or introducing new dedicated circuits for the sake of symbolic computation (Kriete, Noelle, Cohen, \& O’Reilly, 2013; van der Velde, \& de Kamps, 2006).  We will refer to models that adopt some or all of these solutions as the “symbolic connectionist approach” (Holyoak, 1991). Central to the symbolic approach is the claim that conventional connectionist models will fail on symbolic reasoning tasks that humans can perform.

Here we show that conventional connectionist systems can support at least some forms of symbolic processing when trained to output symbolic structures. We achieve this by introducing the vector approach to representing symbols (VARS) that encodes complex symbolic structures of varying complexity as a static numeric vector at the output layer. We show that VARS representations can be learned and that this enables conventional neural networks to achieve combinatorial generalization, a core capacity of symbolic processing.  It is important to emphasize that we do not take our findings to rule out symbolic connectionist architectures -- there may well be functional and biological pressures that lead the brain to adopt special mechanisms devoted to symbolic computation.  But our findings do undermine one of the motivations for this approach - that dedicated mechanisms are necessary to support tasks that require combinatorial generalization.

The structure of the paper is as follows. We start by briefly reviewing the limited successes of conventional networks in modeling tasks that require combinatorial generalization.  We then describe VARS and demonstrate how it can be used to represent symbolic structures in connectionist terms and report two simulation studies showing that neural network models trained to output VARS alongside conventional output representations are able to support an impressive degree of combinatorial generalization in short term memory and visual reasoning tasks.  Importantly, not only do the VARs output representations themselves support combinatorial generalization, but so do the conventional output codes when trained in parallel with the VARS representations. By contrast, the same conventional output codes fail to support combinatorial generalization when the task to output VARS representations is omitted. We argue that our approach does a better job than others existing symbolic and non-symbolic models, and highlights the importance of training conventional neural networks on tasks requiring the explicit representation of symbols.

\section{Review of previous studies assessing combinatorial generalization in conventional connectionist architectures.}
Fodor and Pylyshyn (1988) provided an early seminal criticism of the conventional connectionist framework. They argued that a wide range of cognitive capacities rest on the fact that the mind is compositional, with a small set of context-independent elements (e.g., words, units of meaning) used to productively compose more complex representations in limitless ways. On their view, conventional connectionist models that fail to build in mechanisms to explicitly code for the compositional structure of cognition are doomed to fail in combinatorial generalization tasks in which networks are required to produce novel outputs based on novel combinations of familiar symbols.  For example, after training a network on the symbols John’, ‘Mary’, ‘loves’, and the relation ‘loves (John, Mary)’ a conventional network would not be able to output the relation  ‘loves (Mary, John)’ in response to any query.  Combinatorial generalization is at the heart of what Fodor and Pylyshyn call the systematicity and productivity of thought.

In subsequent debates surrounding this issue, a number of authors highlighted the generalization capacities of conventional connectionist models (e.g., McClelland et al., 2010) and others have highlighted their limitations (Marcus, 1998).  But the conclusions one should draw   with regards to Fodor and Pylyshyn’s critique are far from obvious for a number of reasons.  In some cases the apparent success of a model is not relevant because the model was not actually tested in a condition that required combinatorial generalization (Botvinick \& Plaut, 2006; O’Reilly, 2001; see below for more details). In other cases, the failure of a model is taken as a virtue as it is thought to mirror limited human performance (e.g., Thomas \& McClelland, 2008).  But the more basic problem in reaching any strong conclusion is that the failure of a given conventional connectionist model does not provide a demonstration that all such models will fail. Indeed, many of the early failures of conventional connectionist models that have been used to motivate symbolic models were carried out prior to the development of the more versatile modeling tools and much larger data sets used today.  It is therefore important to assess whether current conventional networks can support combinatorial generalization in tasks that humans can straightforwardly perform. If these models succeed, then symbolic models cannot be motivated on the basis of computational necessity. 

As we summarize next, both earlier and current conventional networks show limited capacity to support combinatorial generalization.  We briefly review these limitations in the domain of short-term memory and visual reasoning tasks, the two domains that we test our VARS model.

\subsection{Combinatorial generalization in sequence learning tasks}

An example of a failure to support combinatorial generalization in an earlier generation of connectionist models was reported by Bowers et al. (2009ab) in the context of modeling short-term memory. Botvinick and Plaut (2006a) had developed simple recurrent model of immediate serial recall that could correctly repeated a sequence of 6 letters approximately 50\% of the time (a level of performance that matches human performance). In addition to accounting for a range of empirical phenomena, the authors had emphasized is that their model could support widespread generalization in that it could recall sequences of letters it had never been trained on.  However, the model was only tested on a limited form of generalization in which each letter was trained in each position within a list.  When Bowers et al. (2009a,b) excluded specific letters in specific positions during training (e.g., the letter A was trained in all positions apart from position 1) and then included them in that position at test, the model did poorly, highlighting the model’s failure in combinatorial generalization. Similar limitations in related networks were observed by other authors (van der Velde, van der Voort van der Kleij, \& de Kamps, 2004; Kriete et al. , 2013).

Do these findings pose a challenge to the conventional connectionist approach to explaining human cognition? Botvinick and Plaut (2009b) defended this approach by arguing that the restricted generalization was a strength based on their claim that humans would also fail under similar training conditions.  Alternatively, the conventional approach might be supported by noting that Bowers et al. (2009ab) only observed limited performance with a simple recurrent network.  More recent and powerful recurrent networks that include long short term memory (LSTM) circuits (Hochreiter \& Schmidhuber, 1997) might well overcome these.  However, both of these lines of defence are difficult to maintain given similar findings have been reported by Lake \& Baroni (2018) using state-of-the-art recurrent networks trained on tasks that humans can transparently perform.  They trained LSTM models to translate a series of commands to a series of actions when the commands were composed of actions (e.g., RUN, WALK) and modifiers of the actions (LEFT, TWICE).  The model was unable to perform the correct series of actions if the model had not been trained an all the relevant combinations of actions and modifiers.  For instance, if the model had never been trained on LEFT-RUN it could not perform the appropriate action despite being trained on LEFT and RUN in other combinations (e.g., LEFT-WALK, TWICE-RUN).  

Still, there are some successes of generalization in sequence learning tasks that appear to require some degree of combinatorial generalization.  For example, Gulordava et al. (2018) trained conventional RNNs on to predict long-distance number agreement in various constructions in four different languages (e.g., predict the verb in: “The girl the boys like: IS or ARE?).  The model was trained on a corpera of text in each language, and critically, succeeded at near human levels not only when tested on sentences composed of meaningful sentences (where predictions might be based on learned semantic or distributional/frequency-based information rather than abstract syntactic knowledge), but also on nonsense sentences that are grammatical but completely meaningless (motivated by the classic sentence by Chomsky: “Colorless green ideas sleep furiously”).  The authors took these findings provide tentative support for the claim that RNNs can construct some abstract grammatical representations.  Nevertheless, when more challenging forms of combinatorial generalization are required, current state-of-the art conventional connectionist systems continue to struggle, as detailed next.

\subsection{Recent explorations of generalization using conventional neural networks in the domain of visual relational reasoning}

Barrett et al. (2018) assessed the capacity of various networks to perform abstract reasoning problems analogous to the Raven-style Progressive Matrices, a well-known human IQ test.  In this task a panel of images are presented that vary according to a rule such as “progression”  (e.g., in a panel of 3 x 3 images in which there is an increasing number of items per image along the first two columns), and the model is trained to select the image that satisfies this rule in order to complete the third column of images (select the target image that has more items). The authors assessed various forms of generalization, including combinatorial generalization (e.g., puzzles in which the progression relation was only encountered when applied to the color of lines and then tested when the progression was applied to the size of shapes). Several state-of-the-art neural network models  were tested and they all performed poorly in the generalization conditions.  The authors were able to improve performance somewhat by adding a “Relation Network” module specifically designed to improve relational reasoning, and more relevant to the current paper, further still by augmenting the training procedure so that the model outputted “meta-targets” that specified the relevant dimensions for correctly responding.  That is, the model was trained not only to select the correct image but also the reason why the image was the correct answer.  Nevertheless, the modified model with the augmented training still performed “strikingly poorly” in the conditions that most relied on combinatorial generalization. 

In a closely related paper, Hill et al. (2019) tested the ability of conventional neural networks to perform analogical reasoning on a set of visual problems.  In this case, the model was presented with a “source” set of three images that shared a given relation (e.g., the number of items in each image increased by one) and a “target” set of two images along with a set of images, only one of which shared the same underlying relation (e.g., progression).  The task of the model was to select the correct image.  Again, the models were tested across a range of conditions, including conditions that required combinatorial generalization, such as (again) applying a familiar relation to new domains.  The authors used a conventional convolutional neural network that provided input to a recurrent layer without any special mechanisms for relational reasoning and found that the type of training had a significant impact on the model’s performance.  When trained in a standard manner in which the model was trained to discriminate the target from foil images that different from one another in various ways the model performed poorly in the combinatorial generalization condition. The important finding, however, was that the model did better in some (but not all) conditions that required combinatorial generation when the training foils were carefully selected so that the model was forced to learn to encode the relevant relations.  The fact that performance continued to be poor in conditions requiring extrapolation highlights how difficult generalization outside the training space can be, but at the same time, the improved performance with carefully crafted training foils suggests that conventional connectionist models can be more successful in such tasks than critics often assume.  The benefits of manipulating the pressure to learn to encode relations for the sake of combinatorial generalization (e.g., Barrett et al.,  2018; Hill et al., 2019) provides the motivation of our approach which we now introduce. 

\section{The vector approach to representing symbols}
The goal of the current paper is to further explore  whether neural network architectures without dedicated mechanisms for symbolic processing can solve the combinatorial generalization problem if they are pushed to learn to explicitly represent symbols. To this end, we train neural network models on two separate tasks -  a main task that answers a query in a standard format (a ‘one hot’ encoding of of the answer) and a secondary task that outputs a symbolic encoding of the problem at hand.  We consider not only whether the model is successful in outputting symbolic representations, but also, whether training on this secondary task leads to success on the main task (the standard one-hot encoding output that typically does not support combinatorial generalisation).

\begin{figure}
	\centering
	\includegraphics{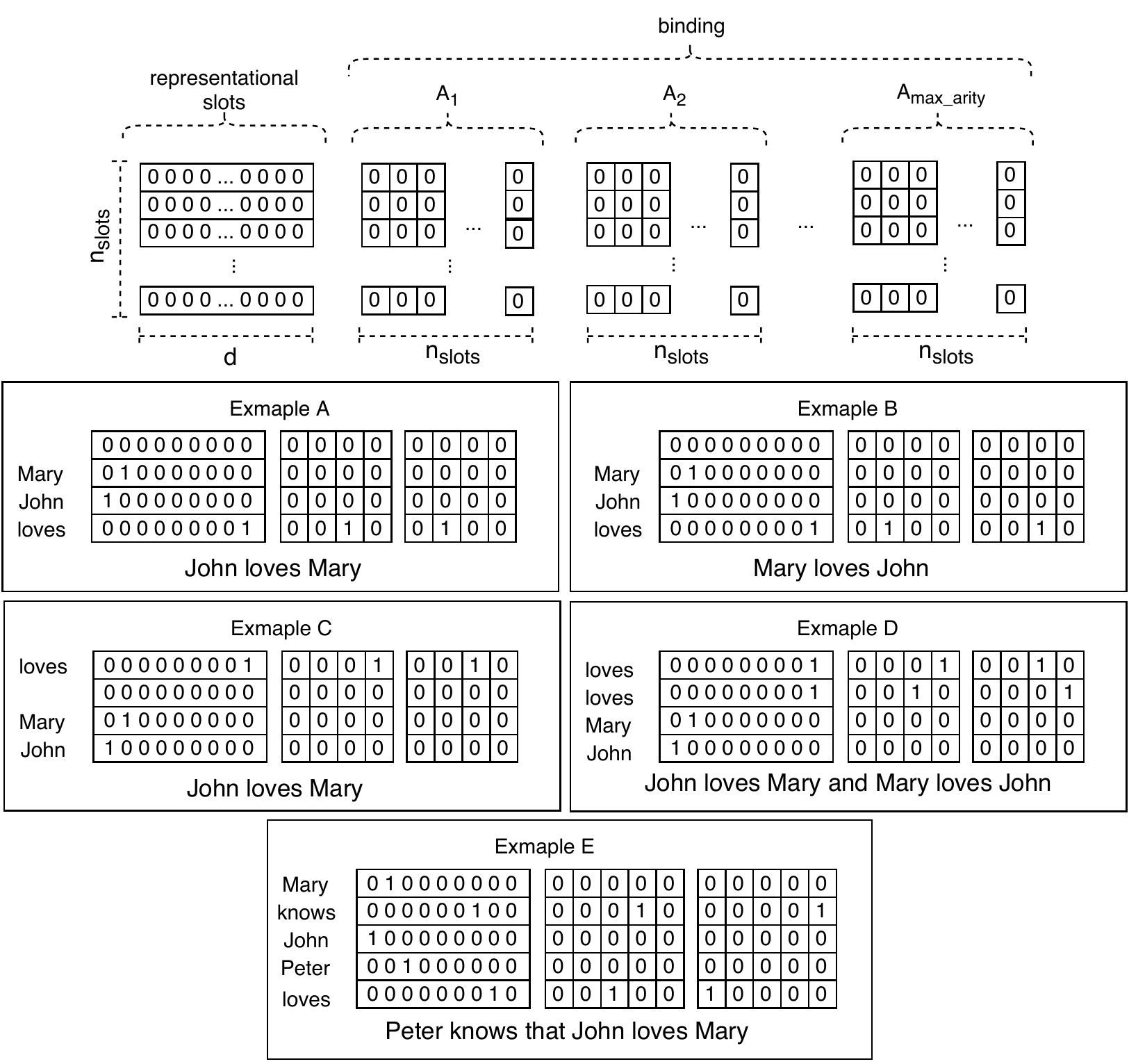}
	\caption{
		The vector approach to representing symbols (VARS). Top: the structure of VARS representations. The number of symbols that can be represented at the same time is limited by the number of representational slots (n\textsubscript{slots}). The maximum arity of the symbols which can be represented is also limited. The symbols are represented by vectors with fixed dimension d, which can be either localist or distributed. For each argument position, there is a separate n\textsubscript{slots} x n\textsubscript{slots} binding matrix (A\textsubscript{1}, A\textsubscript{2}, … A\textsubscript{max\_arity}). Five examples of VARS representations are given. Examples A and B demonstrate how the roles of the “loves” relation are bound to the corresponding fillers. In example A the first argument of “loves” is bound to “John” and the second one is bound to “Mary”. In Example B, the first argument of “loves” is bound to “Mary” by activating the second unit in the fourth row of A1 (“Mary” is represented at the second slot) and the second argument is bound to “John”. Examples A and C represent the same information, although symbols have been located in different slots. Example D demonstrates how two instances of the same type (“‘loves”) are represented. Example E shows the representation of a second order relation (the second argument of “knows” is bound to another relation - “loves”). More examples are available at \url{https://vankov.github.io/combgenvars}
	}
	\label{fig:fig1}
\end{figure}

In order to train on the secondary task  we needed a way to represent symbolic structures as numeric vectors that can be used at the output of standard connectionist models. There have already been several proposals for constructing such representations, including tensor products (Smolensky, 1990) and holographically reduced representations (Plate, 1995)  that are discussed (and criticized) by Hummel (2011). However, there have been no demonstrations so far that conventional neural networks can learn such representations in order to achieve combinatorial generation (but see Schlag \& Schmidhuber, 2018, for a trainable recurrent neural network architecture specifically designed for processing tensor products and Eliasmith, 2013, for a spiking neural network capable of learning holographically reduced representations). We propose a novel way of representing symbolic structures of varying complexity using numeric vectors, which we believe is more suitable for training conventional neural networks on symbolic tasks  - the vector approach to representing symbols or VARS (Figure {\ref{fig:fig1}}).  Our choice to use VARS doesn’t rule out other approaches - a systematic evaluation of the alternative ways to represent symbols in connectionist terms is outside the scope of the current paper.

The main assumptions of VARS are twofold. First, we assume that the meaning of symbols (i.e. the representation of an item independent of its relation to other items) can be encoded at multiple spatial locations within a VARS vector (in either localist or distributed manner). We will refer to these locations as representational slots. The allocation of symbols per slots is arbitrary which means that a symbol can be represented at any slot without affecting its interpretation (Figure {\ref{fig:fig1}}, example A and C). The arbitrary allocation of symbols to slots does not imply that neural representations of symbols can move freely around as the contents of memory cells can move in a computer system. Instead, we assume that there exist redundant representations of symbols which are functionally equivalent, i.e. activating any of them encodes the same knowledge (there are multiple ways to represent the same thing).  In order to achieve such functional equivalence, a system has to be able to represent each symbol at each representational slot and keep the contents of the slot independent of the slot identity (i.e. the representation of “John” should not depend on whether it is activated in slot 1 or slot 2). For models trained to output VARS representations, this can be achieved by learning to encode the symbols at different  slots across trials, while making sure that the allocation of symbols to slots is independent of the information which has to be represented. In this way it is assured that the system will treat the knowledge represented by a symbol independently of the slot which the symbol is allocated to in a given trial. We show how this can be implemented in neural network models in the subsequent simulations.

Being able to encode a symbol at several representational slots allows to represent multiple instances of the same type (see example D in Figure {\ref{fig:fig1}). Moreover, the address of a representational slot, which we will refer to as a “token”, can be used to bind the argument of a predicate to its corresponding filler. The second main principle of VARS is that binding information is represented explicitly and separately from the representation of the meaning of symbols which allows to bind the arguments of predicate symbol to any other symbol, thus ensuring role-filler independence. For each argument position a, there is a separate n\textsubscript{slots} x n\textsubscript{slots} binding matrix A\textsubscript{a}, where n\textsubscript{slots} is the number of representational slots (Figure {\ref{fig:fig1}, top). Binding the n-th argument of the predicate represented at slot i to the symbol at slot j is encoded by activating the unit at the i-th row and the j-th column of A\textsubscript{n}. For example, to represent the binary relation “John loves Mary” (example A in Figure {\ref{fig:fig1}), one would need at least three representational slots and two binding matrixes - A\textsubscript{1} and A\textsubscript{2}, one for each argument position. If “John” is represented at slot 3, “Mary” at slot 2 and “loves” at slot 4, then binding the first argument of “loves” to “John” is implemented by activating the third unit (because “John” is at slot 3) of the fourth row (“loves” is at slot 4) of A\textsubscript{1} (because it represents the binding of the first argument position). Accordingly, activating the second unit of the fourth row of A\textsubscript{2} binds the second argument of “loves” to “Mary”.

The complexity of symbolic structures that can be encoded using VARS is constrained by two parameters: the number of addressable representational slots and the maximum arity of the predicate symbols. The ability of VARS to support role-filler independence and its inherent capacity limits make it a plausible account of how symbolic knowledge is represented in the human mind. However, in this work we use VARS only as a computational pressure to train artificial neural networks to encode knowledge symbolically and we therefore refrain from discussing its cognitive plausibility. 

Using VARS to represent symbolic knowledge bears resemblance to other approaches which use space in order to enable encoding of multiple instances of the same type and role-filler bindings (Marcus, 2001; Kriete et al, 2013; Bowman \& Wyble, 2007; Swan \& Wyble, 2014; van der Velde, \& de Kamps, 2006).  However none of these methods result in fixed size vector representations which can be used to train a conventional neural network architecture. The idea of VARS is also similar to the semantic pointers approach (Eliasmith, 2010).

\section{Simulations}

\subsection{Simulation 1}
In this simulation we replicate a combinatorial generalization task in the context of short-term memory as developed by Kriete et al. (2013).  The model was given a sequence of role-fillers pairs and then cued to recall the filler that was associated with a given role (e.g., after encoding e sequence DOG-SUBJECT, EAT-VERB, STEAK-PATIENT the model was cued to recall the filler EAT when probed with the role VERB). In the combinatorial generalization condition, one of the fillers was never paired with a specific role during training (for example, the filler DOG was never presented in the role of SUBJECT, in any sentence). The authors reported that a simple recurrent network failed on this task whereas a network with an architecture specifically designed to support symbolic processing was successful. 

In order to test whether a recurrent neural network with a conventional architecture can also achieve combinatorial generalization in this task we contrasted the performance of two models (Figure {\ref{fig:fig2}). Both of them included the same conventional long short term memory architecture and were trained to solve the same task, but only one was also trained to output a VARS representation of the structural information provided during the encoding phase (Figure {\ref{fig:fig3}). In this way, the model had to solve two tasks in parallel - the main task which required to output the filler corresponding to the requested role and the VARS representation of the three encoded symbols and their relationship. Performance in combinatorial generalization was measured in both tasks: in the main task this was simply checking whether the correct filler was at the output and in the VARS task we checked whether the slots have been filled and bound correctly. In order to train the model on two tasks, we defined the loss function as a sum of the error on the main task and the VARS task. More details about the simulation are provided in appendix \ref{appendix:a}.

The results of the simulation clearly show that a neural network model without dedicated mechanisms for symbolic processing is much better at combinatorial generalization when pressure to represent knowledge symbolically is enforced (Table {\ref{tab:table1}}). Importantly, the model trained with VARS outputs not only managed to output correct VARS representations of untrained role-filler binding over 90\% of the time, but performed over twice as well on the main task (74\%) compared to the network without VARS (30\%). Performance of the model on the main task in the VARS condition was comparable to the Kriete et al. (2013) model that included specialized mechanisms to support symbolic computations. This finding suggests that training a neural network model to explicitly output symbols qualitatively changes the nature of its internal representations, allowing it to solve problems it otherwise fails on. 

\begin{figure}
	\centering
	\includegraphics{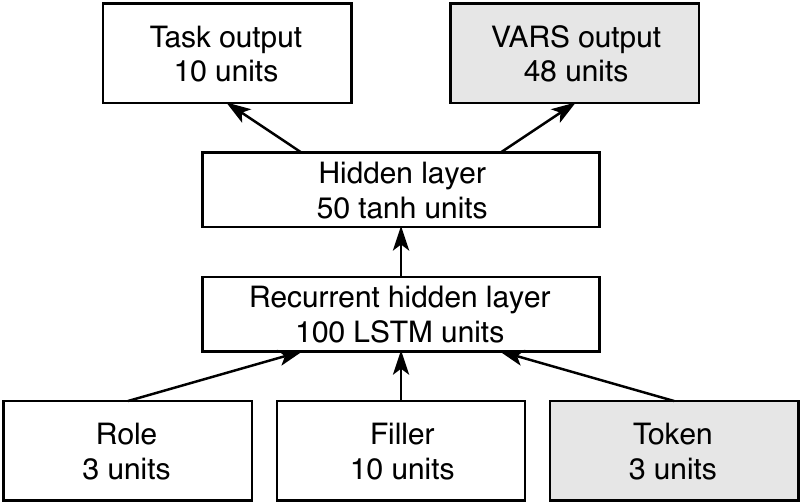}
	\caption{Model architecture in Simulation 1. Two grey components were used only when the model was trained to output VARS representations. The randomly generated token was used to assign the current symbol to the corresponding slot in the VARS output.}
	\label{fig:fig2}
\end{figure}

The results of the simulation clearly show that a neural network model without dedicated mechanisms for symbolic processing is much better at combinatorial generalization when pressure to represent knowledge symbolically is enforced (Table \ref{tab:table1}). Importantly, the model trained with VARS outputs not only managed to output correct VARS representations of untrained role-filler binding over 90\% of the time, but performed over twice as well on the main task (74\%) compared to the network without VARS (30\%).  Performance of the model model on the main task was comparable to the Kriete et al. (2013) model that included special circuit to support symbolic computations. This finding suggests that training a neural network model to explicitly output symbols qualitatively changes the nature of its internal representations, allowing it to solve problems it otherwise fails on.

\begin{figure}
	\centering
	\includegraphics[width=\linewidth]{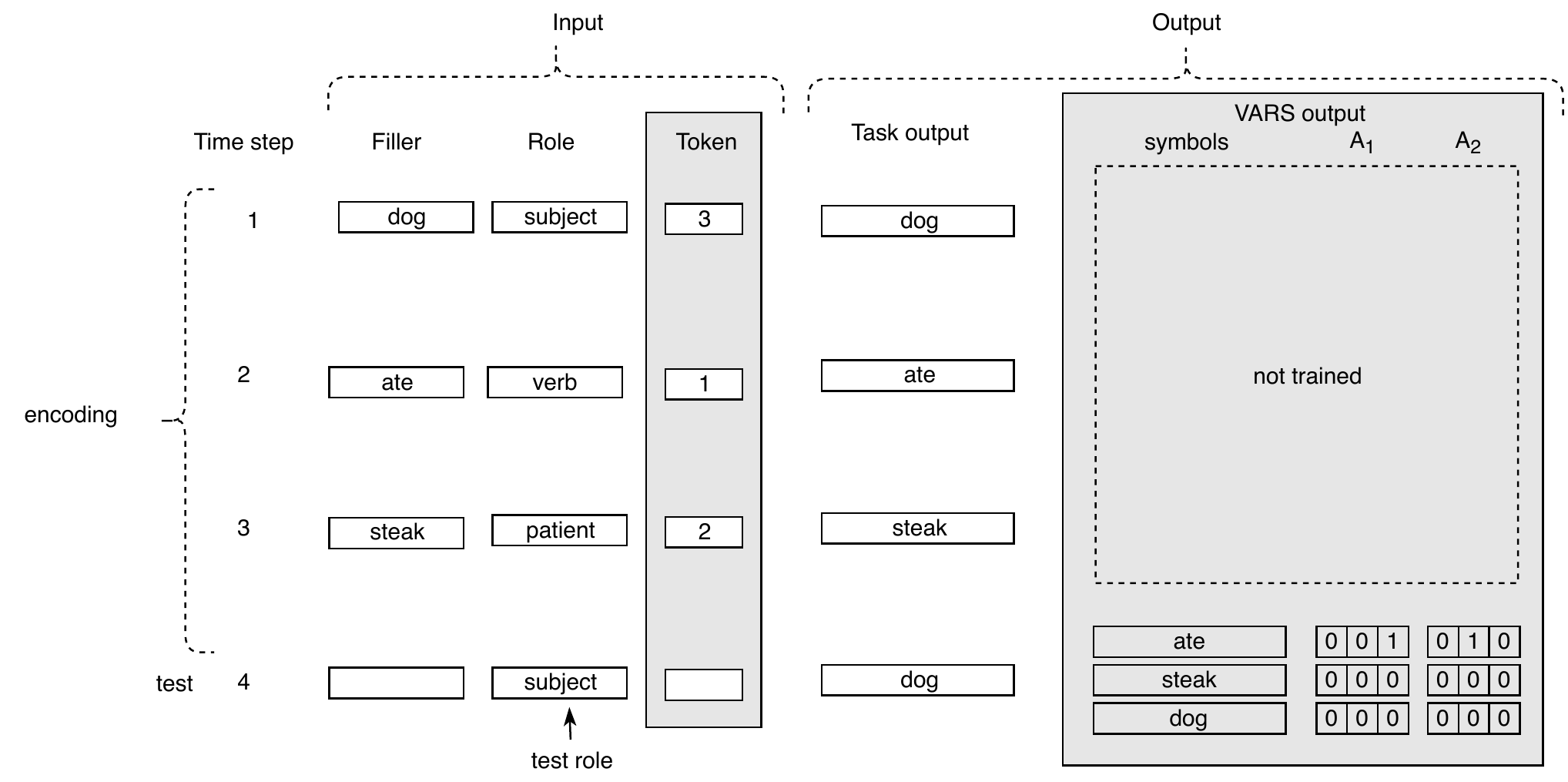}
	\caption{ Description of a sample trial in Simulation 1. During encoding, the model was presented with a series of three role-filler pairs (subject - dog, verb - ate, patient - steak). At the last (fourth) time step, the model was only presented with a test role (in the example above: subject) and it had to output the corresponding filler which was associated to it (dog). When the model was trained to produce VARS representations, the role-filler pairs were accompanied by a random permutation of tokens during encoding which determined the allocation of symbols to representational slots (for example, the fact that dog was paired to token 3 meant that dog has to be represented in the third representational slot). In this way, the random token input ensured that each filler has been allocated to each slot during training.  The verb was treated as a binary relation and its arguments (i.e. the relational roles subject and patient) were bound to the corresponding fillers (in this example, subject to dog and patient to steak). Note, the VARS output was only trained in time step 4, which means that no error was computed during the encoding stage (the ‘not trained’ area).}
	\label{fig:fig3}
\end{figure}

\begin{table}
	\caption{Combinatorial generalization mean accuracy rates in Simulations 1 and 2. In both simulations, the models achieved combinatorial generalization only when trained to explicitly represent symbolic structures. The standard deviation is shown in parentheses.}
	\centering
	\begin{tabular}{clll}

		\multicolumn{4}{r}{Combinatorial generalization accuracy (SD)}                   \\
\cmidrule(r){2-4}
		& Model & Main Task & VARS task \\
		\midrule
		Simulation 1 & LSTM & 0.30 (0.15) & n/a \\
		& LSTM + VARS & 0.74 (0.18) & 0.93 (0.10) \\		
		\midrule		
		Simulation 2 & CNN & 0.29 (0.05) & n/a \\
		& Pre-trained VGG 16 & 0.24 (0.05) & n/a \\		
		& CNN + VARS, no binding & 0.34 (0.11) & 1.00 (0.00) \\				
		& CNN + VARS & 0.99 (0.01) & 0.99 (0.01) \\		
		
		\bottomrule
	\end{tabular}
	\label{tab:table1}
\end{table}

\subsection{Simulation 2}
The goal of our second simulation is to confirm and further extend our finding that conventional neural architectures can achieve combinatorial generalization when pressed to encode knowledge symbolically. Here we assessed whether training a feed-forward convolutional neural network on VARS representations improves combinatorial generalization in a visual reasoning task. 

\begin{figure}
	\centering
	\includegraphics[width=\linewidth]{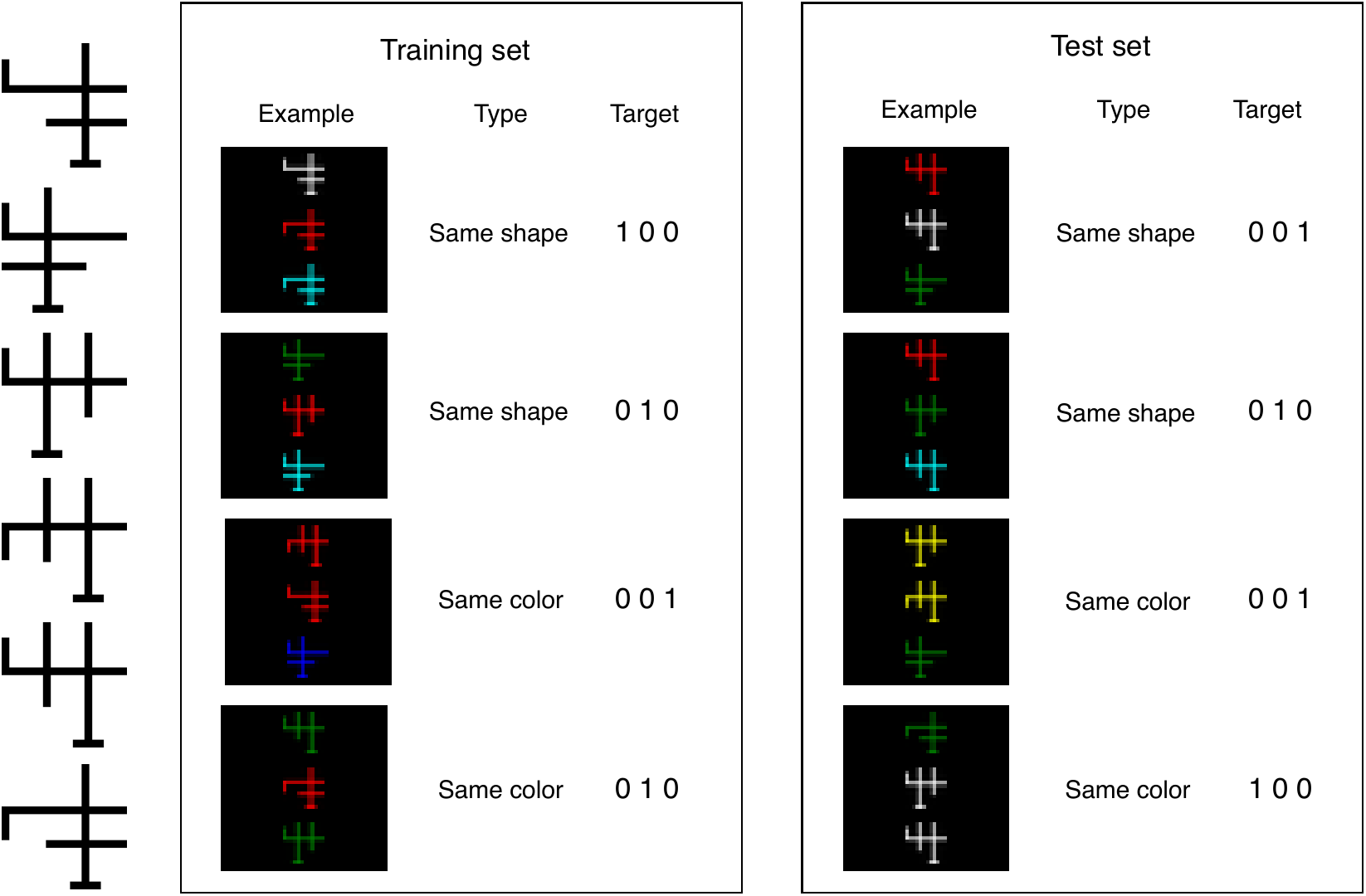}
	\caption{Images used in Simulation 2. The six object shapes are presented on the left. The training set consisted of triplet of objects, arranged in vertical order, such that exactly two of the objects shared a single feature (either same color or same shape). The task of the model was to output the position of the ‘odd’ object by turning the corresponding bit in the target on. For example, in the first example from the training set presented above, the red and the cyan objects have the same shape, so the model has to output the position of the white object (1 0 0). In all of the test examples, the model had to report the position of the green object, which never happened during training position. Note that green object did appear during training, but they were never in the ‘odd’ role (see the second and the fourth training examples).}
	\label{fig:fig4}
\end{figure}

In each trial, the model was presented with three objects and had to choose which one of them was “the odd man out”. Each object had three features: position (top, middle or bottom), shape (one out of six)  and color (one out of six). In each triplet of objects, exactly two of the objects had either the same color or the same shape, but not both (Figure {\ref{fig:fig4}). The task of the model was to output the position of the “odd” object, i.e. the one which shared neither shape nor color with the other two. In order to test combinatorial generalization in this task, we excluded all examples in which the “odd” object was green from the training set and tested the model on these examples. In other words, during training the model never had to report the position of a green object. Green objects did appeared in the training set, but never in the “odd” role. The arbitrary allocation of symbols to representational slots, needed in order to make sure that each symbol can be represented at each slot, was implemented by feeding a random sequence of tokens to the fully connected layers of the network (Figure \ref{fig:fig5} and Figure \ref{fig:fig6}).

\begin{figure}
	\centering
	\includegraphics[width=\linewidth]{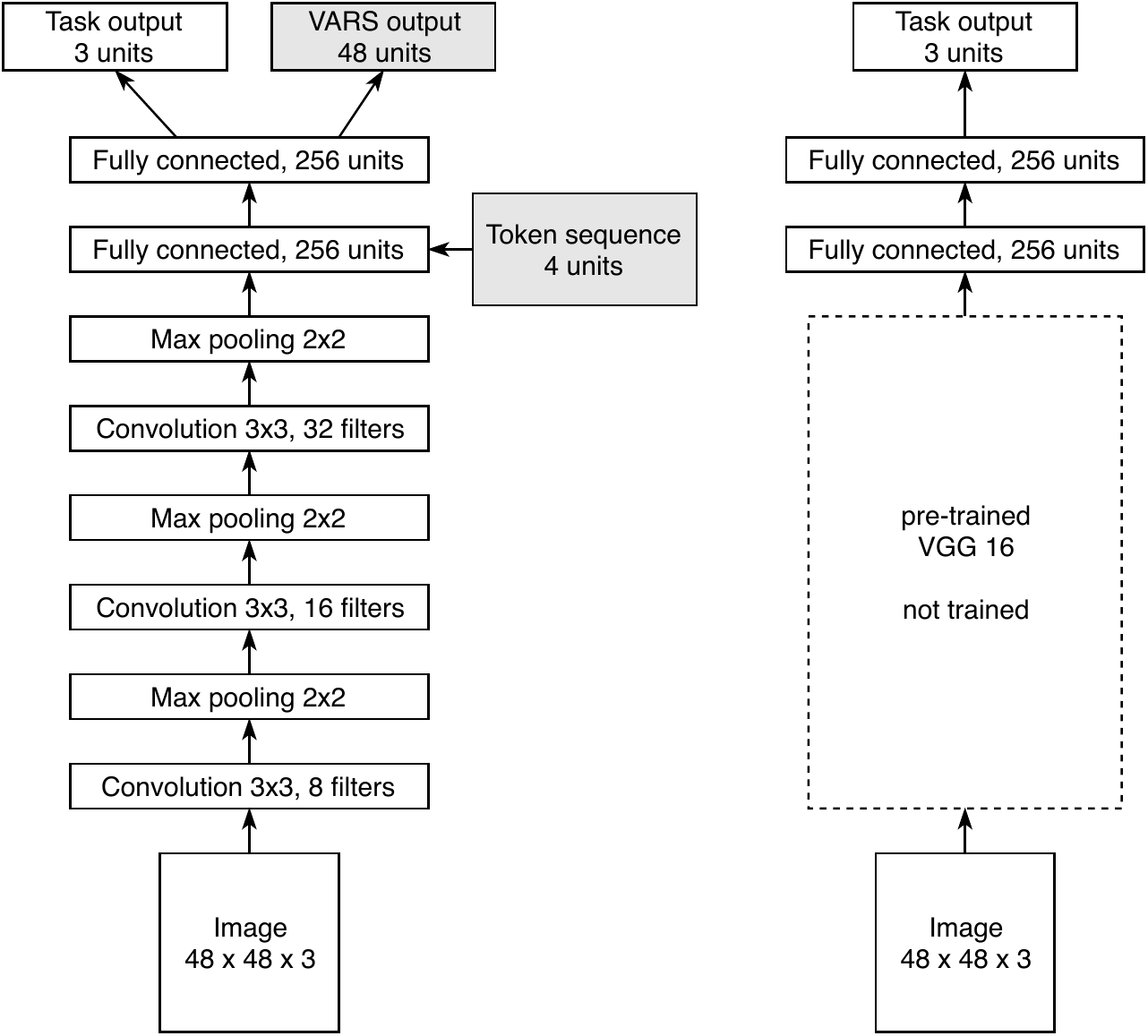}
	\caption{
		Model architectures in Simulation 2. Left: CNN model with three convolutional layers. The components in the grey boxes were only used in the conditions with VARS pressure. Right: pre-trained VGG 16 model with two fully connected layers added at the top. The convolutional layers of VGG 16 were not trained in this simulation. More details about the model are available in appendix \ref{appendix:b}.	
	}
	\label{fig:fig5}
\end{figure}

In order to assess the ability of neural networks to solve the ‘odd man out’ problem with or without the pressure to represent knowledge symbolically we constructed a convolution neural network model displayed in Figure  \ref{fig:fig5}. To make sure that a failure in this task can not be attributed to the details of our custom CNN architecture, we also tested VGG 16 (Simonyan, \& Zisserman, 2015) - a state-of-the-art model of visual object recognition, which has been pre-trained on the ImageNet dataset. The pressure to encode knowledge symbolically was implemented by making our CNN model output VARS parallel to the main task (Figure  \ref{fig:fig6}). Just as in Simulation 1, the loss function the of model trained on VARS was a sum of the error on the main task and the VARS task (more details about the training procedure are available in the supplementary materials) The VARS representation contained the following symbols: X, Y, Z, different-from(X, (Y, Z)), where X, Y, Z belonged to the set (‘top’, ‘middle’ and ‘bottom’).  In other words, the VARS output represented information such as “the middle object is different from the top and from the bottom one”. We also trained the CNN model on VARS representations without binding information (i.e. without the ‘different-from’ symbol) in order to make sure that binding is essential for combinatorial generalization. 

\begin{figure}
	\centering
	\includegraphics[width=\linewidth]{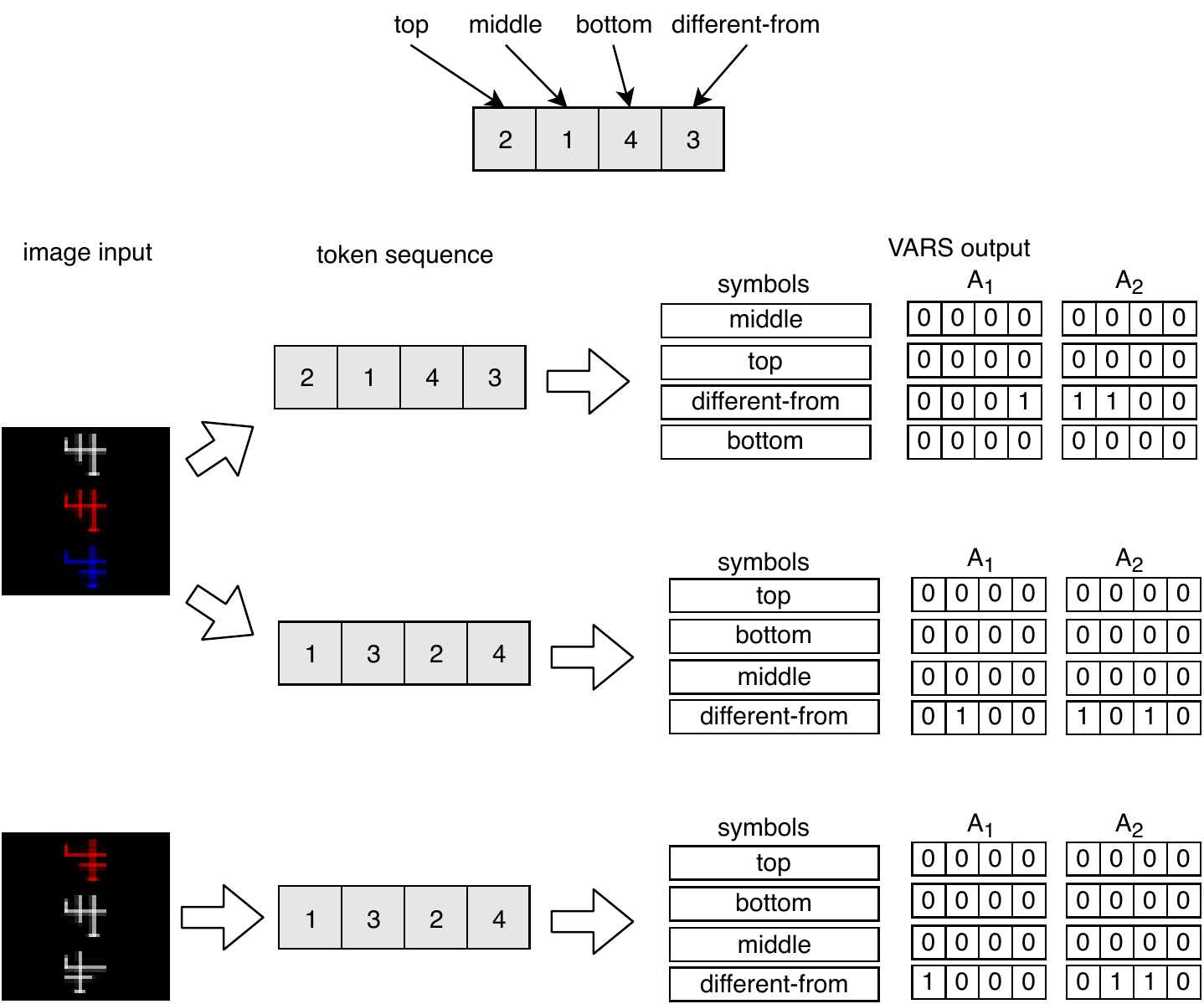}
	\caption{
Encoding VARS representations in Simulation 2. Top - the random sequence of tokens fed to the fully connected layers of the CNN model determined the allocation of symbols to representational slots. In this example, “top” should go to the second slot, “middle” to the first one, “bottom” to the fourth and “different-from” to the third. The first two examples in the bottom pane show how the same image can be encoded in different ways, depending on the order of tokens. The difference between the second and the third examples is only in the binding information, although the odd objects are at different positions, which demonstrate the importance of binding information in order to solve the problem.
	}
	\label{fig:fig6}
\end{figure}

The results of the simulation clearly show that the pressure to encode knowledge symbolically enables it to solve the combinatorial generalization problem (Table {\ref{tab:table1}}). The inability of the model using VARS targets with no binding to account for combinatorial generalization suggests that encoding symbolic structure is indeed what drives the model improvement in generalization and not some idiosyncratic effect of forcing the network to represent the objects at different slots in the VARS output. On the other hand, the failure of the pre-trained VGG 16 model suggests that the difficulty of combinatorial generalization in this task can’t be attributed to the relatively limited visual experience our CNN model was exposed to.

\section{Discussion}
The two simulations above show that combinatorial generalization can be greatly improved in conventional neural networks trained on simple short-term memory and visual reasoning tasks.  In order to achieve this we introduced the vector approach to representing symbols (VARS) that encodes symbolic structures of varying complexity as a static numeric vector in the output layer. The output VARS representations led the models to learn internal representations that better supported combinatorial generalization, not only in the VARS output codes, but also in the main task that used standard ‘one hot’ encoding output units.  This suggests that one of the hallmarks of symbolic computation can be performed without adding new special purpose mechanisms or processes that are often claimed to be necessary.

The current findings are consistent with some recent work that also observed improved combinatorial generalization in conventional connectionist models trained in specific ways designed to improve symbolic computations.  This includes training on carefully tailored training sets that force networks to induce relational representations in order to succeed (Hill et al., 2019), or including output codes that explicitly code for the relevant dimensions of the input patterns, so-called ‘meta-targets’ (Barrett et al., 2018).  The latter finding is similar to our own, although we believe using  VARS is more promising as it allows to train the network on a broader variety of tasks requiring combinatorial generalization, as well as other forms of symbolic processing.

It is important to be clear what we take our contribution to be.  We are not claiming that VARS are necessarily the best way to make conventional networks learn internal representations capable of supporting combinatorial generalization.  There may be other approaches that are as good or better in achieving this goal.  But we are claiming that our findings provide the best evidence to date that conventional neural networks can learn to support combinatorial generalization, and this challenges one of the main motivations for introducing special mechanisms and processes in symbolic networks. 

We are also not claiming that our findings rule out symbolic theories of the mind.  Despite our findings and other recent research (Barrett et al., 2018; Gulordava et al., 2018; Hill et al., 2019), it is still far from clear whether mainstream connectionism can address all the concerns that Fodor and Pylyshyn (1988) raised so long ago. There could be other cases of combinatorial generalizations which can’t be accounted for by purely connectionist models, even when pressures to encode knowledge symbolically are applied. Moreover, there are other and more sophisticated manifestations of symbolic thought, such as language and analogy-making, which still constitute a serious challenge to sub-symbolic approaches. We look forward to more research pushing the boundaries of what neural networks can do without implementing dedicated mechanisms for symbolic processing. And even if such models are shown to support human-level generalization, this alone does not rule out symbolic models.  But then, the debate will not be about whether conventional neural networks are capable of symbolic computation, but about what kind of models provide the best account of human learning and performance.

\section*{Acknowledgements}
Ivan Vankov has been supported by the European Office for Aerospace Research and Development (grant FA9550-15-1-0510). Jeffrey Bowers is grateful for support from the ERC Advanced Grant 741134 (M and M).  

\section*{References}

\bibliographystyle{unsrt}  
%\bibliography{references}  %%% Remove comment to use the external .bib file (using bibtex).
%%% and comment out the ``thebibliography'' section.

\newblock
Antonoglou, I., Fidjeland, A. K., Wierstra, D., King, H., Bellemare, M. G., Legg, S., … Mnih, V. (2015). Human-level control through deep reinforcement learning. {\em Nature, 518} (7540), 529–533.

\newblock
Barrett, D. G., Hill, F., Santoro, A., Morcos, A. S., \& Lillicrap, T. (2018). Measuring abstract reasoning in neural networks. {\em arXiv preprint arXiv:1807.04225}.

\newblock
Battaglia, P. W., Hamrick, J. B., Bapst, V., Sanchez-Gonzalez, A., Zambaldi, V., Malinowski, M., ... \& Gulcehre, C. (2018). Relational inductive biases, deep learning, and graph networks. {\em arXiv preprint arXiv:1806.01261}.

\newblock
Bowers, J.S. (2017). Parallel Distributed Processing Theory in the Age of Deep Networks. {\em Trends in Cognitive Science, 21}, 950-961.

\newblock
Bowers, J.S., Damian, M.F., \& Davis, C.J. (2009a). A fundamental limitation of the conjunctive codes learned in PDP models of cognition-Comments on Botvinick and Plaut. {\em Psychological Review, 116}, 986-995.

\newblock
Bowers, J.S., Damian, M.F.,\& Davis, C.J. (2009b).  More problems with Botvinick and Plaut’s (2006) PDP model of short-term memory. {\em Psychological Review, 116}, 995-997.

\newblock
Bowman, H., \& Wyble, B. (2007). The simultaneous type, serial token model of temporal attention and working memory. {\em Psychological Review, 114}, 1, 38–70.

\newblock
Cooper, R. C., \& Shallice, T. (2006). Hierarchical schemas and goals in the control of sequential behavior. {\em Psychological Review, 113}, 887–916

\newblock
Davis, C. J. (2010). The spatial coding model of visual word identification. {\em Psychological review, 117}, 3, 713-758

\newblock
Doumas, L. A. A., Hummel, J. E., \& Sandhofer, C. M. (2008). A theory of the discovery and predication of relational concepts. {\em Psychological Review}, 115,1-43.

\newblock
Eliasmith, C. (2013). {\em How to build a brain: A neural architecture for biological cognition}. Oxford University Press.

\newblock
Fodor, J. A., \& Pylyshyn, Z. W. (1988). Connectionism and cognitive architecture: A critical analysis. {\em Cognition, 28}(1–2), 3–71. 

\newblock
Gatys. L, Ecker. A., Bethge, M. (2015). A Neural Algorithm of Artistic Style. {\em arXiv preprint arXiv:1508.06576.}

\newblock
Graves, A., Mohamed, A.-R., \& Hinton, G, (2013). Speech recognition with deep recurrent neural networks. In {\em Proceedings of IEEE international conference on acoustics, speech and signal processing}, 6645–6649.

\newblock
Hill, F., Santoro, A., Barrett, D. G., Morcos, A. S., \& Lillicrap, T. (2019). Learning to Make Analogies by Contrasting Abstract Relational Structure. {\em arXiv preprint arXiv:1902.00120.}

\newblock
Hochreiter, S., \& Schmidhuber, J. (1997). Long Short-Term Memory. {\em Neural Computation, 9}(8), 1735–1780. https://doi.org/10.1162/neco.1997.9.8.1735

\newblock
Holyoak, K. J. (1991). Symbolic connectionism: Toward third-generation theories of expertise. In A. Ericsson \& J. Smith (Eds.), {\em Toward a general theory of expertise: Prospects and limits} (pp. 301-355). Cambridge, UK: Cambridge University Press. 

\newblock
Holyoak, K. J., \& Hummel, J. E. (2000). The proper treatment of symbols in a connectionist architecture. In E. Dietrich \& A. B. Markman (Eds.), {\em Cognitive dynamics: Conceptual and representational change in humans and machines} (pp. 229-263). Mahwah, NJ, US: Lawrence Erlbaum Associates Publishers.

\newblock
Holyoak, K., \& Hummel, J. (2003). A symbolic-connectionist theory of relational inference and generalization. {\em Psychological Review, 110}, 2, 220–264.  

\newblock
Hummel, J. E., \& Biederman, I. (1992). Dynamic binding in a neural network for shape recognition. {\em Psychological review, 99}(3), 480.

\newblock
Hummel, J. E., \& Holyoak, K. J. (1997). Distributed representations of structure: A theory of analogical access and mapping. {\em Psychological Review, 104}, 427-466. 

\newblock
Hummel, J. E. (2011). Getting symbols out of a neural architecture. {\em Connection Science, 23}, 109-118.

\newblock
Kokinov, B. \& Petrov, A. A. (2001). Integrating memory and reasoning in analogy-making: The AMBR model. In D. Gentner, K. Holyoak, \& B. Kokinov (Eds.), {\em The analogical mind: Perspectives from cognitive science} (pp. 59-124). Cambridge, MA: MIT Press. 

\newblock
Kriete, T., Noelle, D. C., Cohen, J. D., \& O’Reilly, R. C. (2013). Indirection and symbol-like processing in the prefrontal cortex and basal ganglia. {\em Proceedings of the National Academy of Sciences, 110}(41), 16390-16395.

\newblock
LeCun, Y., Bengio, Y., \& Hinton, G. (2015). {\em Deep learning. Nature, 521} (7553), 436-444.

\newblock
Marcus, G. F. (1998). Rethinking eliminative connectionism. {\em Cognitive psychology, 37}(3), 243-282.

\newblock
Marcus, G. (2001). {\em The algebraic mind: Integrating connectionism and cognitive science.} Cambridge, MA: MIT Press.

\newblock
Marcus, G. (2018). Deep Learning: A Critical Appraisal. {\em arXiv preprint arXiv:1801.00631}

\newblock
McClelland, J. L., Botvinick, M. M., Noelle, D. C., Plaut, D. C., Rogers, T. T., Seidenberg, M. S., \& Smith, L. B. (2010). Letting structure emerge: connectionist and dynamical systems approaches to cognition. {\em Trends in cognitive sciences, 14}(8), 348-356.

\newblock
O’Reilly, R. C. (2001). Generalization in interactive networks: The benefits of inhibitory competition and Hebbian learning. {\em Neural Computation, 13}(6) , 1199–1242

\newblock
Pinker, S., \& Prince, A. (1988). On language and connectionism: Analysis of a parallel distributed processing model of language acquisition. {\em Cognition, 28}(1–2), 73–193. 

\newblock
Plate, T. A. (1995). Holographic reduced representations. {\em IEEE Transactions on Neural Network, 6}(3), 623–641.

\newblock
Silver, D., Huang, A., Maddison, C. J., Guez, A., Sifre, L., van den Driessche, G., Schrittwieser, J., Antonoglou, I., Panneershelvam, V., Lanctot, M., Dieleman, S., Grewe, D., Nham, J., Kalchbrenner, N., Sutskever, I., Lillicrap, T., Leach, M., Kavukcuoglu, K., Graepel, T., \& Hassabis, D. (2016). Mastering the game of go with deep neural networks and tree search. {\em Nature, 529} (7587), 484-489.

\newblock
Schlag, I., Schmidhuber, J. (2018). Learning to Reason with Third Order Tensor Products. In {\em Advances in Neural Information Processing Systems}, 10003–10014.

\newblock
Simonyan, K., \& Zisserman, A. (2015). Very Deep Convolutional Networks for Large-Scale Image Recognition. In {\em Proceedings of 3rd International Conference on Learning Representations (ICLR 2015)}.

\newblock
Smolensky, P. (1990). Tensor product variable binding and the representation of symbolic structures in connectionist systems. {\em Artificial Intelligence, 46}(1-2):159–216, 1990.

\newblock
Swan, G., \& Wyble, B. (2014). The binding pool: A model of shared neural resources for distinct items in visual working memory. {\em Attention, Perception, and Psychophysics, 76}(7), 2136–2157. 

\newblock
Thomas, M. S. C., \& McClelland, J. L. (2008). Connectionist models of cognition. In R. Sun (Ed.), {\em The Cambridge handbook of computational psychology}, (pp.23–58). New York, NY: Cambridge University Press.

\newblock
van der Velde, F., van der Voort van der Kleij, G. T., \& de Kamps, M. (2004). Lack of combinatorial productivity in language processing with simple recurrent networks. {\em Connection Science, 16}(1), 21–46. 

\newblock
van der Velde, F., \& de Kamps, M. (2006). Neural blackboard architectures of combinatorial structures in cognition. {\em Behavioural and Brain Sciences, 29}(1), 37-70. 

\newblock
Wu, Y., Schuster, M., Chen, Z., Le, Q. V., Norouzi, M., Macherey, W., … Dean, J. (2016). Google's Neural Machine Translation System: Bridging the Gap between Human and Machine Translation.{\em arXiv preprint arXiv:1609.08144}

\begin{appendices}
	\section{Simulation 1}
	\label{appendix:a}	
	\subsection*{Data sets}
	The training set was constructed by using a vocabulary of 3 role and 10 filler items (Table {\ref{tab:sables1}}). Roles and fillers were one-hot encoded and the input the model at each time step was a concatenation of the representation of a filler and its corresponding role (Table {\ref{tab:sables1}}, right). All three roles were used within each trial. The order of roles was also varied. Thus, the total number of possible role-filler bindings to encode was 6000. In order to test combinatorial generalization, we excluded all examples in which a particular filler was assigned to a particular role (e. g. filler \#1 to role \#1) from training set, resulting in a 5400 training examples. The set consisted of the remaining 600 trials which contained the untrained role-filler binding.
	
	\setcounter{table}{0}
	\renewcommand{\thetable}{A\arabic{table}}
	\begin{table}[htb]
		\caption{List of filler and roles used in Simulation 1. Note that the labeling of the fillers is arbitrary - each filler could be paired to each role.}
		
		\centering
		\begin{tabular}{crcrcr}
			\multicolumn{2}{c}{fillers} &  \multicolumn{2}{c}{roles} & \multicolumn{2}{c}{example input patterns}\\
			\midrule
			dog & 0 0 0 0 0 0 0 0 0 1 & SUBJECT & 0 0 1 & dog - SUBJECT & 0 0 0 0 0 0 0 0 0 1 1 0 0 \\
			cat & 0 0 0 0 0 0 0 0 1 0 & VERB & 0 1 0 & dog - VERB & 0 0 0 0 0 0 0 0 0 1 0 1 0 \\		
			eat & 0 0 0 0 0 0 0 1 0 0 & PATIENT & 1 0 0 & dog - PATIENT & 0 0 0 0 0 0 0 0 0 1 0 0 1 \\
			steak & 0 0 0 0 0 0 1 0 0 0 & {} & {} & cat - SUBJECT & 0 0 0 0 0 0 0 0 1 0 1 0 0 \\
			fish & 0 0 0 0 0 1 0 0 0 0 & {} & {} & fish -VERB & 0 0 0 0 0 1 0 0 0 0 0 1 0 \\
			take & 0 0 0 0 1 0 0 0 0 0 & {} & {} & eat - PATIENT & 0 0 0 0 0 0 0 1 0 0 0 0 1 \\
			get & 0 0 0 1 0 0 0 0 0 0 & {} & {} & {} & {} \\
			man & 0 0 1 0 0 0 0 0 0 0 & {} & {} & {} & {} \\
			apple & 0 1 0 0 0 0 0 0 0 0 & {} & {} & {} & {} \\
			chase & 1 0 0 0 0 0 0 0 0 0 & {} & {} & {} & {} \\
		\end{tabular}
		\label{tab:sables1}
	\end{table}	
		
	\subsection*{Models}
	The model architecture is shown in Figure {\ref{fig:fig2}} of the main paper. The input unit had 14 units when trained without VARS (10 fillers + 3 roles + 1 unit encoding the test phase) and 17 units when trained on VARS (3 additional additional units indicating the representational slot the current symbol has to be assigned to, see Figure {\ref{fig:fig3}}. The recurrent layer contained standard LSTM units (Hochreiter \& Schmidhuber, 1997), implemented using the Keras deep learning library (Chollet, 2015). No bias was used for the recurrent layer and the activation function was hyperbolic tangent. A fully connected layer with a bias connected the recurrent layer to the output of the model. The activation function of this layer was also hyperbolic tangent. The output of the model was split in two parts - 10 softmax units representing the response to the main task and 48 (3 * 10 for representing the symbols and two binding 3x3 matrices) sigmoid units for the VARS task. 
	
	\subsection*{Training}	
	The model without VARS was trained to minimize the cross entropy of the difference between the output and the target of the main task. When VARS was used, the overall error was the sum of the cross entropy of the main task and the cross entropy of the VARS task. The training proceeded in batches of 50 examples and was organized in epochs of 5000 batches. The optimization algorithm was Adam (Kingma \& Ba, 2015) with a learning rate of 0.001. Both models managed to perfectly fit the examples from the training set.
	
	One of the main assumptions of vector approach to representing symbols is that the representational slots are functionally equivalent, i.e. a DOG activated in slot 1 bears the same interpretation as DOG in slot 3, etc. In order to implement this functional equivalence when training the model to output VARS, we added an additional token input which determined the allocation of symbols to slots. In training each trial, the series of input tokens was randomly permuted which guaranteed that each symbol is trained to be represented at each slot. For example, in Figure {\ref{fig:fig3}} the input to the model consists of the following series of role-filler bindings: (DOG-SUBJECT, ATE-VERB, STEAK-PATIENT) and the token input is (3, 1, 2). Given this input, the model is trained to represent DOG at slot 3, ATE at slot 1 and STEAK at slot 2. A different token input would result in a different allocation of slots. For example, the same series of role-filler bindings (DOG-SUBJECT, ATE-VERB, STEAK-PATIENT) but coupled with another permutation of tokens (1, 3, 2) would result in DOG at slot 1, ATE at slot 3 and STEAK at slot 1. The binding information will also change accordingly.
		
	The performance of the model was assessed in two ways. For the main task, we checked whether the most active unit of the output was the correct one (i.e. the active unit in the target pattern). For the VARS task, we calculated whether the positions of the n most active units in the output match the positions of the active units of the targets, where n is the number of active target units (in this simulation n was always equal to 5 - three target units encoding the symbols in the representational slots and 2 units used for binding).
	
	The model without VARS was trained until the accuracy of the main task in the test set stopped improving for 10 epochs or until the model was trained for 50 epochs, whichever came first. The stopping criterion for the model with VARS was similar, but we used performance on the VARS task in the test set to track training progress. Each simulation was repeated 100 times. The average number of training epochs was 17 (sd = 7.1) and 31 (sd = 12.70)in the conditions without and with VARS, respectively. We also replicated 100 times the simulation with the model without VARS allowing it to run for the maximum of 50 epochs. The results were worse than the ones reported in the main paper, indicating that the amount of training is not responsible for the effect of VARS on combinatorial generalization (mean accuracy 0.27, standard deviation 0.12)
	\subsection*{Source code}		
	\url{https://github.com/vankov/combgenvars/tree/master/simulations/1}
	
	\section{Simulation 2}
	\label{appendix:b}
	\subsection*{Data sets}
	The training set was constructed by generating triplets of objects of varying shape and color, such that exactly two of the objects had the same shape or the same color (but not both).  Three distinct shapes and colors were used. The dimensions of the objects were 12x12 pixels and they were arranged one under the other in the centre of the image (see Figure {\ref{fig:fig4}} in the main text). The size of the images containing the objects was 48x48 pixels. The training set excluded all examples, in which the odd object (the one which was shared neither the same color or the same shape with another one) was green, which resulted in 18000 training examples and 3600 test examples. 
	\subsection*{Models}
	The model architecture is shown in Figure {\ref{fig:fig5}} of the main paper. The dimension of the image input was 48x48x3. The CNN model had three convolutional layers, each having a kernel of size 3x3, rectified linear activation function and a stride of 1. The convolutional layers were followed by 2x2 max pooling transformations. At the end of the hiercharchy, there were two fully connected hidden layers with rectified linear activation functions. The output of the model consisted of three softmax units used to represent the response of the model to the main task (i.e. determining the position of the odd object) and additional 48 sigmoid units when the model was trained to output a VARS representation parallel to the main task. When trained on VARS there an additional input vector of 4 units used to represent the allocation of symbols to slot (Figure {\ref{fig:fig6}}). In each trial, the additional input vector was a random permutation of the numbers from 1 to 4.
	We also trained a second model without VARS, which was based on the VGG 16 architecture (Simonyan \& Zisserman, 2015). That model was already pre-trained on an object recognition task using the ImageNet dataset. We used the convolutional layers of the existing model and added two more fully connected layers on top of them. Training this model affected only the fully connected layers, assuming that the network had already extracted relevant visual features from the much larger training ImageNet dataset.
	\subsection*{Training}		
	The model without VARS was trained to minimize the cross entropy of the difference between the output and the target of the main task. When VARS was used, the overall error was the sum of the cross entropy of the main task and the cross entropy of the VARS task. The training proceeded in batches of 100 examples and was organized in epochs of 500 batches. The optimization algorithm was RMSprop (Tieleman \& Hinton, 2012) with a learning rate of 0.0001. All models were trained to perfection on the training set.
	
	The performance measures were the same as in Simulation 1. For the main task, we checked whether the most active unit of the output was the correct one (i.e. the active unit in the target pattern). All models were trained for 50 epochs and we reported accuracy at the end of training. The simulation was repeated 100 times for each condition (no VARS, VARS with binding, VARS without binding, VGG). We also recorded the maximal performance on the main combinatorial generalization task during training (i.e. at any epoch of training, rather than only at the last one) - the pattern of results was similar no different from the one reported in the main paper (mean accuracy of CNN model: 0.37, VGG 16: 0.34, CNN + VARS without binding: 0.42,  CNN + VARS: 0.99).

	\subsection*{Source code}		
	
	\url{https://github.com/vankov/combgenvars/tree/master/simulations/2}

	\section*{References}
	
	\bibliographystyle{unsrt}  
	%\bibliography{references}  %%% Remove comment to use the external .bib file (using bibtex).
	%%% and comment out the ``thebibliography'' section.
	
	\newblock
	Chollet François. (2015). Keras: The Python Deep Learning library. {\em Keras.Io}. https://doi.org/10.1086/316861
	
	\newblock
	Hochreiter, S., \& Schmidhuber, J. (1997). Long Short-Term Memory. {\em Neural Computation, 9}(8), 1735–1780. https://doi.org/10.1162/neco.1997.9.8.1735
	
	\newblock
	Kingma, D., P. \& Ba. J. (2015). Adam: a Method for Stochastic Optimization. In {\em Proceedings of 3rd International Conference on Learning Representations (ICLR 2015)}.
	
	\newblock
	Simonyan, K. \& Zisserman, A. (2015). Very Deep Convolutional Networks for Large-Scale Image Recognition. In {\em Proceedings of 3rd International Conference on Learning Representations (ICLR 2015)}.
	
	\newblock
	Tieleman, T. \& Hinton, G. (2012) RmsProp: Divide the Gradient by a Running Average of its Recent Magnitude. COURSERA: Neural Networks for Machine Learning.
					
\end{appendices}

\end{document}